\documentclass[conference]{IEEEtran}
\IEEEoverridecommandlockouts

\usepackage{cite}
\usepackage{float}
\usepackage{booktabs}
\usepackage{makecell}
\usepackage{hyperref}
\usepackage{xurl}
\usepackage{subcaption} 
\usepackage{amsmath,amssymb,amsfonts}
\usepackage{algorithmic}
\usepackage{graphicx}
\usepackage{textcomp}
\usepackage{xcolor}
\usepackage{pgfplots}
\usepackage{pgfplotstable}
\pgfplotsset{compat=1.18}
\usepackage{xcolor}
\usepackage{tikz}
\usepackage[T1]{fontenc}
\def\BibTeX{{\rm B\kern-.05em{\sc i\kern-.025em b}\kern-.08em
    T\kern-.1667em\lower.7ex\hbox{E}\kern-.125emX}}
\begin{document}

\title{LAID: Lightweight AI-Generated Image Detection in Spatial and Spectral Domains\\}

 \author{\IEEEauthorblockN{Nicholas Chivaran, Jianbing Ni}
 \IEEEauthorblockA{\textit{Department of Electrical and Computer Engineering, Queen's University} \\
 \textit{ Kingston, ON, Canada}\\
 \{18nc34, jianbing.ni\}@queensu.ca}
 }
\maketitle

\begin{abstract}

The recent proliferation of photorealistic AI-generated images (AIGI) has raised urgent concerns about their potential misuse, particularly on social media platforms. Current state-of-the-art AIGI detection methods typically rely on large, deep neural architectures, creating significant computational barriers to real-time, large-scale deployment on platforms like social media. To challenge this reliance on computationally intensive models, we introduce LAID, the first framework---to our knowledge---that benchmarks and evaluates the detection performance and efficiency of off-the-shelf lightweight neural networks. In this framework, we comprehensively train and evaluate selected models on a representative subset of the GenImage dataset across spatial, spectral, and fusion image domains. Our results demonstrate that lightweight models can achieve competitive accuracy, even under adversarial conditions, while incurring substantially lower memory and computation costs compared to current state-of-the-art methods. This study offers valuable insight into the trade-off between efficiency and performance in AIGI detection and lays a foundation for the development of practical, scalable, and trustworthy detection systems. The source code of LAID can be found at: https://github.com/nchivar/LAID.

\end{abstract}

\section{Introduction}\label{sec:introduction}

The rapid advancement of deep generative models such as Diffusion Models (DMs), Generative Adversarial Networks (GANs), and Variational Autoencoders (VAEs) has enabled the generation of highly photorealistic synthetic imagery. While these models offer significant benefits in areas like creative media, data augmentation, and content generation, they also raise serious concerns about misinformation, deepfakes, and the erosion of digital media authenticity \cite{aigcconcern-lu2023, aigcconcern-peng2024, aigcconcern-misinforeview2024}. These risks are amplified by the staggering volume of images circulated on social media, estimated to be between 1.8 billion to 14 billion images daily across major platforms \cite{smstats-tsvetkova2023, smstats-qut2023, smstats-broz2024photo}. In response to the proliferation of Artificial Intelligence-Generated Content (AIGC) and the explosive growth of image traffic, social media platforms have already begun implementing AIGC detection mechanisms \cite{smstats-clegg2024, smstats-tiktok2023, smstats-xauthenticity2025}. Growing adoption by platform providers highlights the urgent need for detection systems that are robust, accurate, and scalable enough to meet the immense processing demands of real-time social media ecosystems. 

Despite these developments, many state-of-the-art (SOTA) AIGC detection methods remain impractical for large-scale deployment due to their excessive computational overhead. Namely, Artificial Intelligence-Generated Image (AIGI) detection models often depend on deep neural networks with tens to hundreds of millions of parameters and multiple gigaflops (GFLOPs) of computation per sample. This leads to considerable computational costs for both training and inference, limiting adaptability to emerging adversarial techniques and making current SOTA unsuitable for real-time, scalable deployment in social media.

In this paper, we reformulate the AIGI detection problem to overcome scalability concerns by jointly optimizing model performance and computational efficiency. In doing so, we offer practical insight into whether lightweight models can compete with their SOTA counterparts. To this end, we introduce LAID, an AIGI detection framework designed to eliminate computationally expensive preprocessing and complex network pipelines. In LAID we benchmark two existing lightweight model (LiM) paradigms: lightweight convolutional neural networks (LiCNNs) and lightweight vision transformers (LiViTs). We focus on these paradigms as they are among the most prominent and well-studied in computer vision, with extensive open-source implementations that enable accessible detector deployment and minimal data pre-processing.

LAID also leverages both raw spatial (pixel-domain) and simple spectral (frequency-domain) image representations to enable strong discriminative power in LiMs despite their reduced size. This aligns with recent literature that shows that combining these complementary perspectives can reveal distinct detection artifacts \cite{related-guo2025, related-durall2020, related-tomen2021, related-he2021, related-jeong2022}. Furthermore, we rigorously evaluate the adversarial robustness of LiMs to ensure their practical reliability in adversarial settings.

We summarize our contributions as follows:

\begin{enumerate}
    \item \textbf{First lightweight AIGI detection benchmark:} We introduce LAID, the first benchmark of lightweight AIGI detection---to our knowledge---which approaches AIGI detection through the lens of accuracy and scalability.
    
    \item \textbf{Feasibility for real-world LiM deployment:} We show that LiMs can achieve near-perfect detection accuracy while significantly reducing computational and memory costs, making them well-suited for deployment in high-throughput environments such as social media platforms.
    
    \item \textbf{Lightweight robustness under adversarial conditions:} We conduct a thorough evaluation of the adversarial robustness of LiMs, revealing trade-offs between model complexity, detection accuracy, and robustness which are critical in securing real-world detection pipelines.
\end{enumerate}

The remainder of this paper is organized as follows. Section \ref{sec:related} reviews related AIGI detection works and Section \ref{sec:background} provides a background on LiMs and adversarial attacks to contextualize LAID. Section \ref{sec:methodology} outlines our LAID framework, followed by Section \ref{sec:experiments} and Section \ref{sec:results} which discuss our experimental setup and results respectively. Finally, Section \ref{sec:conclustion} concludes the paper and suggests future research directions.

\section{Related Works} \label{sec:related}
We begin our work by reviewing the landscape of AIGI detection methods, categorizing them by detection paradigm and design philosophy.

\subsection{Active vs. Passive AIGI Detection}\label{subsec:related-passive-vs-active}
AIGI detectors can be broadly classified into two categories: active and passive methods \cite{related-deng2024}. Active methods approach detection by intervening during the image generation process where  forensic signals such as watermarks, fingerprints, or metadata are embedded to facilitate subsequent identification \cite{related-deng2024}. Further distinction in active methods is made between non-semantic and semantic schemes, with the latter gaining prominence as a promising direction for robust and content-aware detection \cite{related-Kassis2024}. Passive methods approach detection post-hoc, analyzing  images after they have been generated to detect synthetic content. These approaches rely on techniques such as CNN classification, spectral domain analysis, and the identification of textural or statistical inconsistencies \cite{related-deng2024}.

Guo et al. \cite{related-guo2025} provide a comparative analysis of the two paradigms within their ImageDetectBench benchmark, revealing that watermark-based (active) methods generally outperform passive methods in both clean and adversarial detection scenarios. This finding aligns with a broader trend in AIGI detection literature, where active methods are being increasingly recognized for their robustness and effectiveness. Notable non-semantic active schemes include HiDDeN \cite{related_Zhu2018}, StableSignature \cite{related-fernandez2023}, WOUAF \cite{related-kim2024}, and Gaussian Shading \cite{related-yang2024}, while semantic schemes include Tree-Ring Watermarks (TRW) \cite{related-wen2023}, DeepTag \cite{related-wang2020}, and StegaStamp \cite{related-tancik2020}. Although active techniques offer strong verification guarantees in cooperative scenarios, they typically require modifications image generation pipelines. This limits their applicability in real-world contexts where content origin is unknown, watermarking is not applied, or the watermarking approach is not publicly available. Given our goal of developing a non-invasive, lightweight solution, we focus the rest of our literature review on passive detection works.

\subsection{Passive AIGI Detection Methods}\label{subsec:related-passive}
Recent research in passive AIGI detection encompasses a diverse range of strategies aimed at enhancing detection performance, generalization across models and datasets, and robustness to adversarial attacks.

\subsubsection{Spatial Analysis}\label{subsubsec:related-passive-img}
One of the extensively studied areas in passive AIGI detection is spatial analysis, a paradigm based on the observation that artificially generated images often exhibit pixel-level and textural inconsistencies introduced during the image generation process \cite{related-Rossler2019, related-wang2020-CNNSpot, related-tan2023, related-ojha2023, related-wang2023, related-zhong2024}. By examining these image-domain cues, detectors can more effectively differentiate synthetic content from real images.

Rössler et al. \cite{related-Rossler2019} introduced FaceForensics++, a benchmark and framework that applied XceptionNet to raw facial pixels from GANs, establishing an early baseline for adversarial AIGI detection. Building on this, Wang et al. \cite{related-wang2020-CNNSpot} proposed CNNSpot, which employed a ResNet50 network trained to capture spatial inconsistencies in AIGI. Zhong et al. \cite{related-zhong2024} introduced PatchCraft which demonstrated that inter-pixel correlations between texture-rich and texture-poor image patches reveal a universal fingerprint for synthetic content. Finally, Wang et al. \cite{related-wang2023} presented DIRE, a diffusion-model detector that identifies AIGI using a residual-based reconstruction error derived from Denoising Diffusion Implicit Models (DDIM).

\subsubsection{Spectral Analysis}\label{subsubsec:related-passive-freq}
Another key area of passive AIGI detection is spectral detection which builds on the insight that deep image generation models often introduce frequency-domain artifacts due to architectural limitations and their typical reliance on upsampling operations \cite{related-durall2020, related-tomen2021, related-he2021, related-jeong2022}. These artifacts deviate from the natural frequency distributions of real images, making them effective features for detecting AIGI.

Durall et al. \cite{related-durall2020} were among the first to demonstrate that CNN-based generative models (e.g. GANs) suffer from poor frequency reproduction and distinctive spectral artifacts that can be detected by a support vector machine (SVM). Building on this, Tomen et al. \cite{related-tomen2021} showed that the widespread use of small kernel sizes in CNN-based generators leads to spectral leakage, resulting in both high- and low-frequency distortions. Frank et al. \cite{related-frank2020} used discrete cosine transform (DCT)  features with the GAN fingerprinting proposed in \cite{related-yu2019}. Similarly, Yan et al. \cite{related-yan2025} integrated DCT component analysis with semantic features as the input to a ResNet50 classifier. More recently, Jeong et al. \cite{related-jeong2022} proposed their Bilateral High-Pass Filter (BiHPF) framework, which enhances spatial artifacts in synthetic images using two complementary high-pass frequency-domain filters followed by a ResNet50 classifier.

\subsubsection{Alternative Passive Methods}\label{subsubsec:related-passive-other}
In addition to spatial and spectral approaches, a set of passive AIGI detection methods employ distinct, alternative strategies. Ojha et al. \cite{related-ojha2023} and Sha et al. \cite{related-sha2023} take advantage of the representational power of CLIP \cite{related-radford2021} in their works by using CLIP embeddings as inputs to a classification network. Tan et al. \cite{related-tan2023} developed LGrad, a method that uses the gradients of several popular neural networks (e.g. CLIP \cite{related-radford2021}, InceptionV3 \cite{related-Szegedy2016} and ViT \cite{related-dosovitskiy2020}) as feature inputs to a ResNet-50 classifier. He et al. \cite{related-he2021} proposed a resynthesis-based solution where image inputs are passed through reconstruction networks trained exclusively on real images. Since synthetic images differ from real images distributionally, they are poorly reconstructed, making their reconstructions readily detectable. Recently, Wang et al. \cite{related-wang2021-fakespotter} proposed FakeSpotter, a neuron-coverage technique that uses a ResNet50 classifier to distinguish synthetic images from real images based on inputted VGG-Face \cite{related-VGGFace} neuron activation patterns.

\subsection{Lightweight AIGI Detection}\label{subsubsec:related-lightweight}
While there is extensive literature on various aspects of AIGI detection, relatively few works have focused on lightweight detection explicitly. 
Durall et al. \cite{related-durall2020} evaluated AIGC detection methods using minimal preprocessing and a compact, easily integrable architecture---highlighting a key research gap in lightweight AIGI detection pipelines. Lađević et al. \cite{related-ladevic2024} presented a custom CNN pipeline that achieved 97.32\% accuracy on the CIFAKE dataset. Similarly, Mulki et al. \cite{related-mulki2024} employed a MobileNetV2-based architecture to achieve 77.79\% accuracy on a dataset containing real and synthetic images generated by sources such as Stable Diffusion \cite{related-rombach2022}, Midjourney \cite{related-midjourney}, and DALL·E \cite{related-ramesh2021}.

\section{Background}\label{sec:background}
To provide context for our work in AIGI detection, we briefly introduce modern LiM approaches and standard adversarial attacks in AIGI.

\subsection{Lightweight Models}\label{subsec:background-lightweight}
Lightweight models (LiMs) are architectures developed to remedy the large computational demands of deep learning models. This makes them more suitable for devices with limited resources (e.g. mobile phones) or real-time applications (e.g. social media). Within LiMs two  popular architectural choices are lightweight CNNs (LiCNNs) and lightweight vision transformers LiViTs. LiCNNs achieve efficiency through techniques such as depthwise separable convolutions and optimized network designs. Conversely, LiViTs adapt transformer architectures for vision tasks by incorporating architectural simplifications and hybrid convolution-attention mechanisms. 

\subsection{Adversarial Robustness in AIGI Detection} \label{subsec:related-adversaries}
To evaluate the robustness of AIGI detection models, adversarial attacks are designed to introduce subtle distortions and perturbations that preserve an input's quality but also degrade any model’s predictive performance. While prior research has explored custom, targeted adversarial attacks, our work focuses on evaluating robustness with general attacks commonly encountered in both practical applications and literature \cite{related-frank2020, related-Kassis2024, related-yu2019, related-yan2025}, offering insight into applicable, real-world robustness.

 \begin{figure*}[t]
    \centering
    \includegraphics[width=\textwidth]{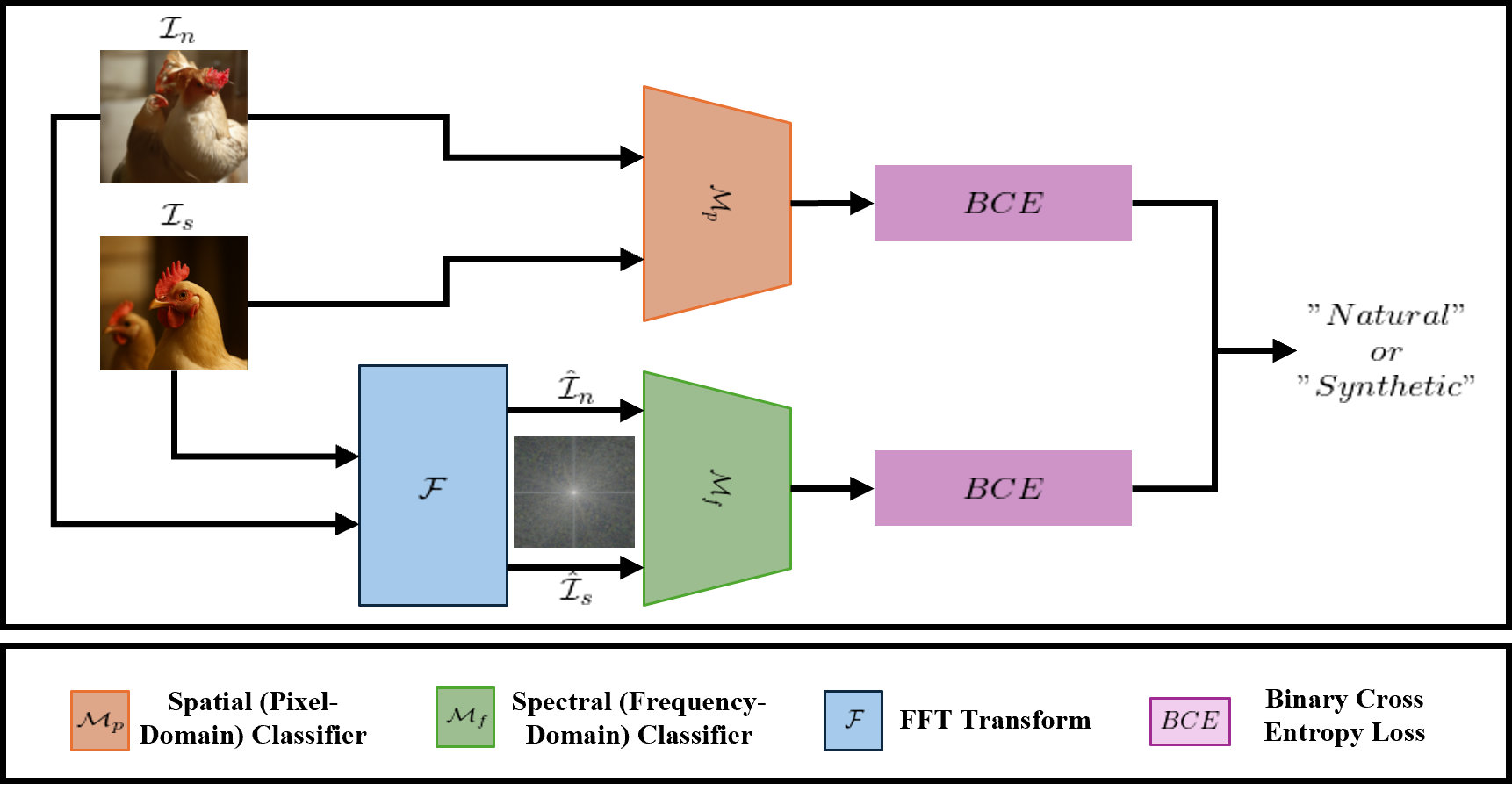}
    \caption{LAID training pipeline. Testing removes binary cross-entropy (BCE) and additionally evaluates input images ($\mathcal{I}_n$, $\mathcal{I}_s)$ that have been perturbed (see Section \ref{subsec:experiments-testing})}
    \label{fig:pipeline}
\end{figure*}

\section{LAID}\label{sec:methodology}
This section outlines the core methods of LAID, including problem formulation, pipeline, and training strategy.

\subsection{Problem Formulation}\label{subsec:methodology-problem}
LAID is an AIGI detection framework that uses LiMs to differentiate between images generated by generative models (synthetic images) and images that have been captured naturally (natural images). Formally, given an input image $\mathcal{I}\in\mathbb{R}^{H \times W \times C}$, let $\mathcal{M}$ be a LiM that learns the mapping $\mathcal{M}_\theta (\mathcal{I})\rightarrow\{0,1\}$ where 0 denotes natural images, 1 denotes synthetic images, and $\theta$ denotes the parameters of $\mathcal{M}$. 

Unlike prior works which solely focused on maximizing detection accuracy, we frame AIGI detection with explicit constraints on detection performance and computational efficiency---aiming to identify LiMs that are practical for deployment on social media platforms. Specifically, we constrain our evaluation to LiMs that satisfy at least one of the following two criteria: (1) Fewer than 10M learnable parameters ($|\theta| < 10^7$) or (2) less than 1 GFLOP per inference operation (\(\text{FLOPs} < 10^9\)). The first constraint limits analysis to models that prioritize memory efficiency, increasing the parallelization potential of model deployment. The second constraint targets computational efficiency, minimizing potential latency for users of social media applications. In addition to these constraints, LAID makes no assumptions about the design or operation of the underlying generation models used to generate AIGI (i.e. a black-box setting).

To enhance the reliability of LAID's evaluated detectors, we leverage both spatial (pixel) and spectral (frequency) domain representations of input images. As such, there are 2 versions of each evaluated model $\mathcal{M}$: one that is trained on spatial domain images ($\mathcal{M}_p$) and another that is trained on spectral domain frequency plots ($\mathcal{M}_f$). The spectral images used for training $\mathcal{M}_f$ are simply zero-centered 2D Fast Fourier Transforms ($\mathcal{F}$) of corresponding spatial images: $\mathcal{\hat{I}} = \mathcal{F}(\mathcal{I})$. It is important to note that $\mathcal{M}_p$ and $\mathcal{M}_f$ are trained independently on images from their respective domains.

\subsection{LAID Pipeline}\label{subsec:methodology-pipeline}
Figure \ref{fig:pipeline} illustrates the LAID pipeline. Given either a natural ($\mathcal{I}_n$) or synthetic ($\mathcal{I}_s$) input image, LAID employs two images representations:  raw spatial-domain images ($\mathcal{{I}}_n$, $\mathcal{{I}}_s$) and their frequency-domain counterparts ($\mathcal{\hat{I}}_n$, $\mathcal{\hat{I}}_s$). The two representations are then passed through their corresponding models (\(\mathcal{M}_p\), \(\mathcal{M}_f\)), training the two architecturally-identical yet separate models where model outputs are generated independently.

During testing, we evaluate models independently under both clean and adversarial settings. We also explore a basic fusion approach in adversarial settings (see Section \ref{subsec:experiments-testing}), facilitating the evaluation of varied deployment constraints for settings with varied computational demands.

\subsection{Training Strategy} \label{subsec:method-training}
We adopt a standardized training procedure where each selected model is initialized using weights ($\theta_{pre}$) pretrained on ImageNet-1K. This allows for spatial-domain priors to be leveraged (see Table \ref{tab:selection}). For baseline methods where pretrained weights are unavailable (see Section \ref{subsec:experiments-selection}), we adjust our initialization approach such that the model parameters for  Lađević et al. \cite{related-ladevic2024} are randomly initialized and SpottingDiffusion \cite{related-mulki2024} uses a pretrained MobileNetV2 backbone with all remaining parameters randomly initialized. After initialization, models are then finetuned for our AIGI detection task using the GenImage dataset (see Section \ref{subsec:experiments-dataset}). 

All models are trained using Binary Cross-Entropy (BCE) loss for 100 epochs with an Adam optimizer ($\text{lr}=1\times10^{-4}$, $\lambda_{wd}=0$). We also use a ReduceLROnPlateau scheduler ($p=5$, $\gamma=0.1$) to accelerate convergence. To regularize training and reduce redundant computation, we apply an early stopping mechanism where training is terminated after 10 consecutive epochs without improvement in validation accuracy. Additionally, we use PyTorch’s automatic mixed precision (AMP) and a batch size of 512. The best-performing model checkpoints---based on validation accuracy---are saved for final testing.

\section{Experimental Setup} \label{sec:experiments}
This section discusses our dataset and its preprocessing, model selection procedure, testing strategy, evaluation metrics, and deployment environment. 

\subsection{Dataset}\label{subsec:experiments-dataset}
We use the GenImage dataset \cite{related-zhu2023}, a large-scale dataset for AIGI detection evaluation, for training and evaluation in LAID. GenImage contains approximately 1.33M natural images and 1.35M synthetic images originating from 8 generative sources: Stable Diffusion 1.4 \cite{related-rombach2022}, Stable Diffusion 1.5 \cite{related-rombach2022}, Midjourney \cite{related-midjourney}, ADM \cite{related-dhariwal2021}, GLIDE \cite{related-nichol2021}, Wukong \cite{related-wukong2022}, VQDM \cite{related-Egiazarian2024}, and BigGAN \cite{related-brock2018}. 

\begin{figure}[htbp]
    \centering
    \caption{Image from spatial dataset and its corresponding zero-centered 2D-FFT spectral representation.}
    \label{fig:conversion}
    \begin{subfigure}[b]{0.45\linewidth}
        \centering
        \includegraphics[width=\linewidth]{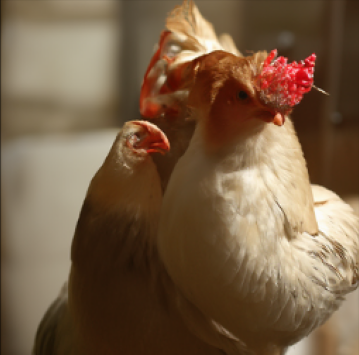}
        \caption{Spatial-Domain Image}
        \label{fig:raw-image}
    \end{subfigure}
    \hfill
    \begin{subfigure}[b]{0.45\linewidth}
        \centering
        \includegraphics[width=\linewidth]{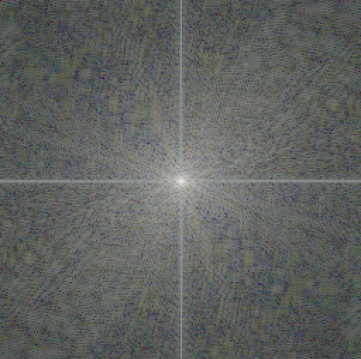}
        \caption{Frequency-Domain Plot}
        \label{fig:spectral-image}
    \end{subfigure}
\end{figure}

Given that GenImage only contains train and test sets, we first evenly split GenImage's original test set into validation and test data to allow for performance monitoring on unseen data during training.
To make training feasible, we construct our spatial dataset by randomly subsampling 100,000 training images, 16,000 validation images, and 16,000 testing images across class-generator pairs (e.g. 6,250 natural samples and 6,250 synthetic samples from the MidJourney subset of GenImage). Random sampling is used to mitigate selection and sampling bias. Additionally, as discussed in Section \ref{subsec:methodology-problem}, we construct a spectral dataset from our spatial dataset by applying a zero-centered 2D-FFT to each spatial image (see Figure \ref{fig:conversion}), preserving distribution consistency between datasets. 

\begin{table*}[t]
\caption{Lightweight models ranked and selected via efficiency score. Value in parentheses of the \textit{Rank} column reflects the original ranking of models amongst model variants filtered by parameter ($|\theta| < 10^7$) and FLOP ($\text{FLOPs} < 10^9$) constraints.}
\centering
\resizebox{\textwidth}{!}{%
\begin{tabular}{llccccc}
\toprule
\textbf{Rank} & \textbf{Model} & \textbf{Initialization Weights ($\theta_{pre}$)} & \textbf{Top-1 Acc. (\%) ~$\uparrow$} & \textbf{\# Parameters (M) ~$\downarrow$} & \textbf{GFLOPs ~$\downarrow$} & \textbf{Efficiency Score ($\mathcal{E}$)~$\uparrow$} \\
\midrule
1 (1) & ShuffleNet & X0\_5\_Weights.IMAGENET1K\_V1
& 60.55 & 1.4 & 0.04 & 0.806 \\

2 (2) & EdgeNeXt & EdgeNeXt-XXS 
& 71.20 & 1.3 & 0.26 & 0.671 \\

3 (3) & MobileNetV3 & Small\_Weights.IMAGENET1K\_V1 
& 67.67 & 2.5 & 0.06 & 0.669 \\

4 (4) & MobileViT & MobileViT-XXS
& 69.00 & 1.3 & 0.40 & 0.645 \\

5 (5) & MobileViTV2 & MobileViT-v2-0.5
& 70.20 & 1.4 & 0.50 & 0.631 \\

6 (6) & MNASNet & 0\_5\_Weights.IMAGENET1K\_V1
& 67.73 & 2.2 & 0.10 & 0.619 \\

7 (8) & SqueezeNet & 1\_1\_Weights.IMAGENET1K\_V1
& 58.18 & 1.2 & 0.35 & 0.607 \\

8 (16) & MobileNetV2 & Weights.IMAGENET1K\_V2 
& 72.15 & 3.5 & 0.30 & 0.526 \\

9 (19) & FastViT & FastViT-T8
& 75.60 & 3.6 & 0.70 & 0.524 \\

10 (20) & RegNet & Y\_400MF\_Weights.IMAGENET1K\_V2
& 75.80 & 4.3 & 0.40 & 0.523 \\
\bottomrule
\end{tabular}%
}
\label{tab:selection}
\end{table*} 

\subsection{Data Preprocessing}\label{subsec:experiments-preprocessing}
We strictly normalize both spatial and spectral images to the [0, 255] range, resize them to $256\times256 \ \text{pixels}$, and store them as PyTorch tensors. These preprocessing steps reduce model training time while preserving the original spatial and frequency characteristics of the dataset. 

\subsection{Model Selection} \label{subsec:experiments-selection}
To ensure a fair and reproducible framework, we sourced candidate LiMs from the official PyTorch model hub \cite{related-pytorchzoo} and Hugging Face PyTorch image models library \cite{related-timm}. Both repositories provide standardized collections of pretrained models, guaranteeing the consistent initialization of models and eliminating the need for complex implementations. Since both sources contain models that exceed our problem constraints (see Section \ref{subsec:methodology-problem}), we first filter out models that do not meet our parameter and FLOP limits. The only exceptions are for our baseline methods Lađević et al. \cite{related-ladevic2024} and SpottingDiffusion \cite{related-mulki2024}, which were selected as current lightweight SOTA approaches. After constraint filtering, the list of candidate models is: for LCNNs--AlexNet \cite{related-Krizhevsky2012}, DenseNet \cite{related-huang2017}, EfficientNet \cite{related-tan2021}, GoogLeNet \cite{related-szegedy2015}, MnasNet \cite{related-tan2019}, MobileNetV2 \cite{related-sandler2018}, MobileNetV3 \cite{related-howard2019}, RegNet \cite{related-xu2022}, ShuffleNet \cite{related-zhang2018}, SqueezeNet \cite{related-iandola2016}; and for LiViTs--CoaT \cite{related-xu2021}, EdgeNeXt \cite{related-maaz2022}, EfficientFormer \cite{related-li2022}, FastViT \cite{related-vasu2023}, LeViT \cite{related-graham2021}, MobileViT \cite{related-mehta2021}, and MobileViTV2 \cite{related-mehta2022}. 

To select which LiMs we benchmark in LAID alongside SOTA baselines, we propose an efficiency score $\mathcal{E}$ as our selection criterion. The efficiency score of a model quantifies its accuracy and computational efficiency in one metric by combining the model's: (1) top-1 accuracy on the ImageNet1k dataset (to ensure strong spatial-domain understanding), (2) FLOPs per sample (to penalize high computational cost), and (3) number of model parameters (to penalize large model size). Mathematically, $\mathcal{E}$ is computed as:
\begin{equation}
\resizebox{0.91\linewidth}{!}{$
\mathcal{E} = \lambda_1\left(\frac{\text{Acc}_{\text{Top-1}}}{\max(\text{Acc})}\right)
+ \lambda_2\left(\frac{\min(\text{FLOPs})}{\text{FLOPs}}\right)
+ \lambda_3\left(\frac{\min(\text{Params})}{\text{Params}}\right)
$}
\end{equation}

where $\lambda_1$, $\lambda_2$, and $\lambda_3$ control the relative importance of accuracy, computational efficiency, and model size respectively. For our experiments, we select \(\lambda_1 =0.5\), \(\lambda_2 =0.25\), and \(\lambda_3 =0.25\) to balance performance (accuracy) and resource constraints (FLOPS and parameters) equally. 

We rank our constraint-filtered LiCNNs and LiViTs based on their $\mathcal{E}$ and select the top 10 models for evaluation. To avoid architectural redundancy, we only select the highest-ranking model from each parent architecture. For instance, although multiple pretrained ShuffleNet variants \cite{related-zhang2018} appeared amongst top entries, we only evaluate the model using X0\_5\_Weights.IMAGENET1K\_V1 weights. A summary of our selection results can be found in Table \ref{tab:selection}. Note that we modify the final linear layer of selected LiMs and baseline models to ensure they have two outputs, aligning with the binary AIGI detection task.

\begin{table*}[t]
\centering
\caption{Clean testing performance results both spatial and spectral models. Bolded rows indicate the best performing models for each domain.}
\begin{tabular}{c|ccc|ccc}
\toprule
\textbf{Model} & \multicolumn{3}{c|}{\textbf{Spatial Domain}} & \multicolumn{3}{c}{\textbf{Spectral Domain}} \\ \cmidrule(r){2-4} \cmidrule(l){5-7}
& \textbf{Accuracy (\%)~\(\uparrow\)} & \textbf{F1-Score~\(\uparrow\)} & \textbf{AUC-ROC~\(\uparrow\)} & \textbf{Accuracy (\%)~\(\uparrow\)} & \textbf{F1-Score~\(\uparrow\)} & \textbf{AUC-ROC~\(\uparrow\)} \\
\midrule
ShuffleNet          & 92.40 & 0.9229 & 0.9782 & 99.44 & 0.9944 & 0.9983 \\
EdgeNeXt            & 95.64 & 0.9563 & 0.9916 & 99.77 & 0.9977 & 0.9999 \\
MobileNetV3         & 95.19 & 0.9523 & 0.9911 & 99.67 & 0.9968 & 0.9995 \\
MobileViT           & 97.17 & 0.9716 & 0.9943 & 99.89 & 0.9989 & 0.9999 \\
MobileViTV2         & 97.01 & 0.9700 & 0.9962 & 99.50 & 0.9950 & 0.9997 \\
MnasNet             & 78.11 & 0.7837 & 0.8642 & 50.36 & 0.6631 & 0.6892 \\
SqueezeNet          & 92.47 & 0.9245 & 0.9796 & 99.37 & 0.9937 & 0.9996 \\
MobileNetV2         & 97.86 & 0.9786 & 0.9976 & 99.55 & 0.9955 & 0.9997 \\
FastViT             & 98.87 & 0.9887 & 0.9989 & \textbf{99.91} & \textbf{0.9991} & 0.9999 \\
RegNet              & 97.98 & 0.9798 & 0.9973 & 99.89 & 0.9989 & \textbf{1.000} \\
Lađević et al.  
& \textbf{99.22} & \textbf{0.9923} & \textbf{0.9993} & 99.79 & 0.9979 & 0.9998 \\
SpottingDiffusion
& 98.74 & 0.9874 & 0.9982 & 99.61 & 0.9961 & 0.9997 \\ 
\bottomrule
\end{tabular}
\label{tab:clean-perf}
\end{table*} 

\subsection{Testing Strategy}\label{subsec:experiments-testing}
According to Section \ref{subsec:methodology-pipeline}, we formally outline our testing approach in LAID. Our testing protocol assesses both clean performance and adversarial robustness while also exploring the benefits of decision-level fusion to enhance robustness. Models are evaluated independently under two scenarios:
\begin{enumerate}
    \item \textbf{Clean Testing:} Respective spatial and spectral models are evaluated on clean (non-attacked) spatial and spectral images. This can be formulated as:
    \begin{equation}
    {y}_p = M_p(\mathcal{I}), \quad {y}_f = M_f(\mathcal{\hat{I}}),
    \end{equation}
    
    where detection is successful when predicted labels match the ground truth ($\mathcal{G}$) (i.e. \(y_p=\mathcal{G}\) and \(y_f=\mathcal{G}\)).

    \item \textbf{Adversarial Testing:} Adversarial training consists of three evaluation settings: (a) spatial evaluation on perturbed spatial images (b) spectral evaluation using $\mathcal{F}$-transformed perturbed images, and (c) fusion evaluation which combines (a) and (b) through decision-level fusion. These settings can be formulated as:
    \begin{equation} \textbf{(a)}\quad
    {y}_p = M_p(\mathcal{I}_{adv}),
    \end{equation}
    \begin{equation} \textbf{(b)}\quad
    {y}_f = M_p(\mathcal{\hat{I}}_{adv}),
    \end{equation}
    \begin{equation} 
    \textbf{(c)}\quad
    {y}_p = M_p(\mathcal{I}_{adv}), \quad 
    {y}_f = M_p(\mathcal{\hat{I}}_{adv}),
    \end{equation}
    
    where detection is successful when predictions match the ground truth (i.e. \(y_p=\mathcal{G}, y_f=\mathcal{G}\)) for (a) and (b) and when either prediction matches the ground truth (i.e \((y_p = \mathcal{G})\lor(y_f=\mathcal{G})\)) for (c).
\end{enumerate}

Decision-level fusion showed little advantage in clean scenarios, so we chose to only evaluate it in adversarial testing

\subsection{Adversarial Attack Setting}\label{subsec:experiments-attack}

To evaluated selected LiMs in adversarial settings, we subject models to 5 adversarial attacks: 

\begin{itemize}
    \item \textbf{Cropping:} Removes between 5–20\% of the image area and then resizes the image to to $256\times256 \text { pixels}$ to simulate occlusion or framing shifts.
    
    \item \textbf{Blurring:} Applies Gaussian blur with a kernel size randomly selected from \{3, 5, 7, 9\} and standard deviation from \{1.0, 2.0, 3.0, 4.0\} to mimic low-quality optics or post-processing effects.

    \item \textbf{Noising:} Adds Gaussian noise with variance uniformly sampled between 5 and 20 to replicate sensor noise or compression-related distortions.

    \item \textbf{JPEG Compression:} Compresses the image with a JPEG quality factor randomly sampled between 25 and 90 to reflect typical online image degradation.
    
    \item \textbf{Combination:} Applies individual attacks with a 50\% probability to simulate compound perturbations often found in real-world settings.
    
\end{itemize}

These attacks simulate transforms commonly applied by social media platforms to reduce storage demands and accelerate content delivery. Their design and strength are derived from recent AIGI detection literature \cite{related-frank2020, related-Kassis2024, related-yu2019, related-yan2025}.

\subsection{Evaluation Metrics}\label{subsec:experiments-metrics}
To evaluate detection performance in LAID, we employ accuracy, F1-score, and AUC-ROC (Area under the Receiver Operating Characteristic Curve). These three metrics provide a holistic view on how well models address AIGI detection. 

To evaluate the resource efficiency of models in LAID, we use absolute metrics---parameter count and GFLOPs---to capture space and time complexity as well as ratio-based metrics---accuracy per million parameter (Acc/MParams) and accuracy per MFLOP (Acc/MFLOPs)---to assess how efficiently LiMs utilize their resources. 

\subsection{Experimental Environment}\label{subsec:experiments-environment}
Experiments were performed on systems with an NVIDIA A100 GPU, Intel Xeon Platinum 8362, and 128 GB of memory with Python 3.12.4, PyTorch 2.1.2, and CUDA 12.1.


\section{Results and Analysis} \label{sec:results}
In this section, we evaluate selected LiMs in clean and adversarial testing scenarios as well as high-level trends that suggest promise for LiMs in AIGI detection.

\subsection{Clean Testing Performance}\label{subsec:results-clean}
The results in Table \ref{tab:clean-perf} reveal several insights into the detection performance of LiMs across clean image domains.

\subsubsection{Spatial Domain Performance}\label{subsubsec:results-clean-img}
Most LiMs achieve strong spatial detection accuracy ranging between 92\% and 99\%. These results are supported by strong F1-scores and near perfect AUC-ROC values indicating that models are precise and consistent in their detection. Among evaluated LiMs, Ladevíc et al. achieves the highest spatial detection performance, featuring the only accuracy above 99\% (99.22\%). In contrast, the sole exception to the otherwise viable LiMs is MnasNet, which features a significantly lower accuracy (78.11\%), F1-score (0.7837), and AUC-ROC (0.8642). Overall, these findings suggest that LiMs are highly effective for clean AIGI detection---performing on-par with SOTA approaches---and no single model is universally superior in this setting.

\subsubsection{Spectral Domain Performance}\label{subsubsec:results-clean-freq}
Spectrally, almost all LiMs show a significant performance boost over their spatial counterparts, with all models---except MnasNet, which falls to a near-chance accuracy of 50.36\%---achieving over 99\% accuracy as well as near-perfect F1-scores and AUC-ROC values. This consistent improvement suggests that frequency-based features provide a more discriminative representation for clean AIGI detection. Furthermore, frequency-based features narrow the performance gap between smaller and larger LiMs, suggesting that they are simpler and more easily accessible for computationally constrained models. Once again, no single LiM is universally optimal for clean detection, but all evaluated models demonstrate strong and reliable performance.

\begin{figure*}[t]
    \centering
    \caption{Spatial and spectral visualization of dataset image before and after 50\% JPEG compression. Note the minimal visual difference between raw images contrasted with the pronounced transformation of the frequency domain.}
    \label{fig:fft-comparison}

    \begin{subfigure}[b]{0.23\linewidth}
        \centering
        \includegraphics[width=\linewidth]{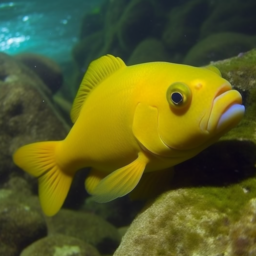}
        \caption{Original Image (Raw)}
    \end{subfigure}
    \hfill
    \begin{subfigure}[b]{0.23\linewidth}
        \centering
        \includegraphics[width=\linewidth]{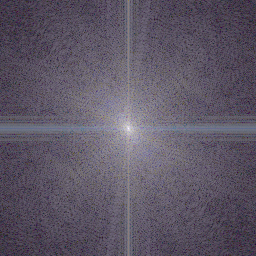}
        \caption{Original Image (FFT)}
    \end{subfigure}
    \hfill
    \begin{subfigure}[b]{0.23\linewidth}
        \centering
        \includegraphics[width=\linewidth]{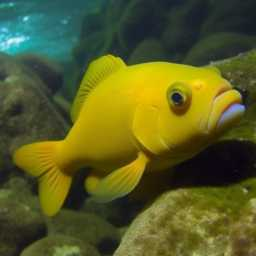}
        \caption{Attacked Image (Raw)}
    \end{subfigure}
    \hfill
    \begin{subfigure}[b]{0.23\linewidth}
        \centering
        \includegraphics[width=\linewidth]{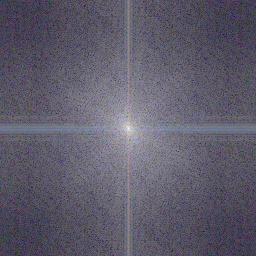}
        \caption{Attacked Image (FFT)}
    \end{subfigure}
\end{figure*}

\subsubsection{Comparative Insights}\label{subsubsec:results-clean-comp}
Across both input domains, evaluated LiMs exhibit consistently high detection performance on clean AIGI data with accuracies surpassing 92\%. These results are reinforced by strong F1-scores and near-perfect AUC-ROC values, demonstrating that LiMs are highly effective for clean AIGI detection. Moreover, findings show that frequency-based input features enhances LiM performance, suggesting that spectral approaches are generally more effective.

When comparing architectural classes, LiViTs generally outperform their LiCNN counterparts in the spatial domain. This suggests that the attention mechanisms and global receptive fields of ViTs better capture the spatial features needed for AIGI detection. Spectrally however, this advantage largely disappears as both LiViTs and LiCNNs exhibit near-identical accuracies, F1-scores, and AUC-ROC values.

\begin{table}[h]
\centering
\caption{Accuracy (\%) across spatial, spectral, and fusion attack settings. Bolded rows indicate the most robust models.}
\resizebox{\linewidth}{!} {%
\begin{tabular}{c|ccccc}
\toprule
\textbf{Model} & \textbf{Crop} & \textbf{Blur} & \textbf{Noise} & \textbf{Compress} & \textbf{Combined} \\
\midrule
\multicolumn{6}{c}{\textbf{Spatial Setting}} \\
\midrule
ShuffleNet          & 80.50 & 52.66 & 79.38 & 82.31 & 68.16 \\
EdgeNeXt            & 80.41 & 53.09 & 77.47 & 78.69 & 67.72 \\
MobileNetV3         & 82.84 & 55.62 & 83.41 & \textbf{84.09} & \textbf{73.44} \\
MobileViT           & 81.50 & 52.12 & 78.75 & 65.06 & 63.53 \\
MobileViTV2         & 82.81 & 52.44 & 80.09 & 62.47 & 64.97 \\
MnasNet             & 69.25 & 51.66 & 76.16 & 74.75 & 61.41 \\
SqueezeNet          & 79.84 & 54.22 & 81.91 & 81.94 & 68.94 \\
MobileNetV2         & 84.03 & 52.72 & 77.44 & 58.44 & 63.56 \\
FastViT             & \textbf{88.75} &\textbf{ 55.47} & 82.44 & 60.94 & 65.09 \\
RegNet              & 83.41 & 52.94 & \textbf{87.09} & 63.44 & 65.66 \\
Lađević et al.      & 80.78 & 51.03 & 82.54 & 54.59 & 59.59 \\ SpottingDiffusion   & 86.12 & 52.78 & 79.16 & 58.81 & 62.59 \\
\midrule
\multicolumn{6}{c}{\textbf{Spectral Setting}} \\
\midrule
ShuffleNet          & 50.16 & 50.09 & 49.97 & 50.03 & 50.12 \\
EdgeNeXt            & 50.09 & 50.09 & 50.00 & 49.97 & 50.06 \\
MobileNetV3         & 50.38 & 50.03 & 50.00 & 49.88 & 50.00 \\
MobileViT           & 50.03 & 50.03 & 50.00 & 49.97 & 50.06 \\
MobileViTV2         & 50.00 & 50.00 & 50.00 & 49.84  & 49.94 \\
MnasNet             & 46.88 & 49.81 & 49.41 & 50.00 & 49.81 \\
SqueezeNet          & 50.06 & 50.00 & 50.00 & 49.78 & 49.94 \\
MobileNetV2         & \textbf{50.81} & \textbf{50.28} & 50.00 & 49.84 & 50.03 \\
FastViT             & 51.06 & 51.06 & 50.00 & 49.50 & \textbf{50.22} \\
RegNet              & 50.06 & 50.00 & \textbf{50.12} & \textbf{50.12} & 49.91\\
Lađević et al.      & 50.00 & 50.19 & 50.00 & 49.91 & 50.00 \\
SpottingDiffusion   & 50.09 & 50.00 & 50.00 & 50.06 & 50.03 \\
\midrule
\multicolumn{6}{c}{\textbf{Fusion Setting}} \\
\midrule
ShuffleNet          & 85.88 & 53.09 & 96.62 & 95.91 & 84.25 \\
EdgeNeXt            & 86.47 & 53.97 & 99.31 & 96.78 & 88.50 \\
MobileNetV3         & 87.38 & 56.09 & 97.50 & 93.88 & 83.19 \\
MobileViT           & 87.59 & 53.22 & 99.25 & 99.50 & 89.97 \\
MobileViTV2         & 87.16 & 53.00 & 96.94  & 99.34 & 89.28 \\
MnasNet             & 75.31 & \textbf{57.47} & 91.75 & 84.28 & 68.44 \\
SqueezeNet          & 85.81 & 55.03 & 97.22 & 96.66 & 84.34 \\
MobileNetV2         & 87.41 & 53.66 & 98.97 & 99.09 & 90.66 \\
FastViT             & 92.56 & 56.50 & 98.31 & 98.34 & 90.91\\
RegNet              & 86.19 & 53.06 & \textbf{99.81} & 99.59 & 88.53 \\
Lađević et al.      & \textbf{98.38} & 55.84 & 98.59 & 99.19 & 90.50 \\
SpottingDiffusion   & 89.62 & 53.69 & 99.72 & \textbf{99.66} & \textbf{91.19} \\
\bottomrule
\end{tabular}%
}
\label{tab:results-adversarial}
\end{table}

\subsection{Adversarial Testing Performance}\label{subsec:results-attack}
Table \ref{tab:results-adversarial} outlines the results of our adversarial testing.

\subsubsection{Spatial Setting}\label{subsubsec:results-attack-spatial}
As shown in Table \ref{tab:results-adversarial}, the adversarial robustness of LiMs varies greatly based on the attack and model. Blurring and compression attacks lead to the greatest performance degradation with accuracies dropping to near-random levels. This effect extends to combination attacks where most models achieve around a 60\% detection accuracy. In contrast, cropping attacks impact performance significantly less, with most models maintaining around an 80\% accuracy. Notably, our compression exposes a clear robustness gap between LiCNNs and LiViTs, where a subset of LiCNNs---ShuffleNet, MobileNetV3, and SqueezeNet---remain resilient against compression (80\%+ accuracy) and all LiViTs experience a sharp accuracy, falling to at best 65.06\%. This disparity suggests that lightweight convolutional architectures may inherently possess greater tolerance to low-frequency distortions, underscoring their practicality over LiViTs in compression-heavy environments such as social media platforms.

\subsubsection{Spectral Setting}\label{subsubsec:results-attack-spectral}
Table \ref{tab:results-adversarial} reveals the universal collapse of spectral LiMs in attacked settings with all attack accuracies near or slightly below random guessing. This vulnerability stems from the frequency domain’s sensitivity to spatial alterations where even minor spatial changes can large spectral distortions as illustrated in Figure \ref{fig:fft-comparison}. These findings emphasize the fragility of spectral features under adversarial conditions, suggesting that spectral LiMs require defenses such as adversarial training to remain adversarially robust. 

\begin{table*}[t]
\centering
\caption{Clean testing efficiency results for spatial and spectral models. Bolded rows indicate the most efficient architecture.}
\begin{tabular}{c|cc|cc}
\toprule
\textbf{Model} 
& \multicolumn{2}{c|}{\textbf{Spatial Domain}} 
& \multicolumn{2}{c}{\textbf{Spectral Domain}} \\
\cmidrule(r){2-3} \cmidrule(r){4-5}
& \makecell{\textbf{Acc./MParams} \\(\%/MParams)~\(\uparrow\)} 
& \makecell{\textbf{Acc./MFLOP} \\(\%/MFLOPs)~\(\uparrow\)} 
& \makecell{\textbf{Acc./MParams} \\(\%/MParams)~\(\uparrow\)} 
& \makecell{\textbf{Acc./MFLOP} \\(\%/MFLOPs)~\(\uparrow\)} \\
\midrule
ShuffleNet          & \textbf{268.72} & \textbf{1.62} & \textbf{289.21} & \textbf{1.75} \\
EdgeNeXt            & 82.66 & 0.37 & 86.23 & 0.39 \\
MobileNetV3         & 62.62 & 1.19 & 65.58 & 1.25 \\
MobileViT           & 102.55 & 0.29 & 105.42 & 0.30 \\
MobileViTV2         & 88.03 & 0.21 & 90.29 & 0.21\\
MnasNet             & 83.09 & 0.52 & 53.57 & 0.33 \\
SqueezeNet          & 127.80 & 0.27 & 137.34 & 0.29 \\
MobileNetV2         & 43.96 & 0.23 & 44.72 & 0.23 \\
FastViT             & 30.54 & 0.14 & 30.87 & 0.14 \\
RegNet              & 25.10 & 0.18 & 25.58 & 0.18 \\
Lađević et al.      & 20.44 & 0.01 & 20.56 & 0.01 \\
SpottingDiffusion   & 4.26 & 0.22 & 4.29 & 0.22 \\
\bottomrule
\end{tabular}
\label{tab:clean-efficiency}
\end{table*}

\subsubsection{Fusion Setting}\label{subsubsec:results-attack-fusion}
Fusion offers a substantial improvement in adversarial robustness for LiMs, achieving higher adversarial accuracy when spatial and spectral inputs are combined. While marginal gains were anticipated---given that our fusion strategy should perform no worse than the better of the two domain models---the significant improvement observed suggests additional underlying factors. 

The adversarial attacks chosen in this work exploit domain-specific vulnerabilities. For example, cropping, blurring, and compression disrupt images at a structural and texture level, degrading the performance of spatial models. In contrast, noising and other spectral attacks target an image's frequency distribution, undermining spectral model performance. As a result, joint consideration of spectral and spatial predictions during fusion allows LiMs to analyze redundant, diverse, and complimentary feature spaces. It is important to recognize that, although fusion-based strategies enhance adversarial robustness, they also incur additional computational cost.

\subsection{LiM Efficiency}\label{subsec:results-eff}
Table \ref{tab:clean-efficiency} shows that ShuffleNet demonstrates the highest resource efficiency, achieving an accuracy of 268\%/MParams and 1.62\%/MFLOP spatially as well as 289\%/MParams and 1.75\%/MFLOP spectrally. These results are expected due to both the relatively narrow performance gape between LiMs as well as ShuffleNet having the smallest computational footprint and one of the smallest memory footprints.

While the selected models balance computational and storage efficiency, the two SOTA baselines exhibit a clear imbalance between them. Lađević et al. has poor computational efficiency due to its considerable requirement of 13.46 GFLOPs per sample. Similarly, SpottingDiffusion has the worst memory efficiency due its immense 23M parameter count. These results reinforce that choosing pretrained LiMs over current SOTA solutions should be the starting point in systems that aim to be resource-conscious.

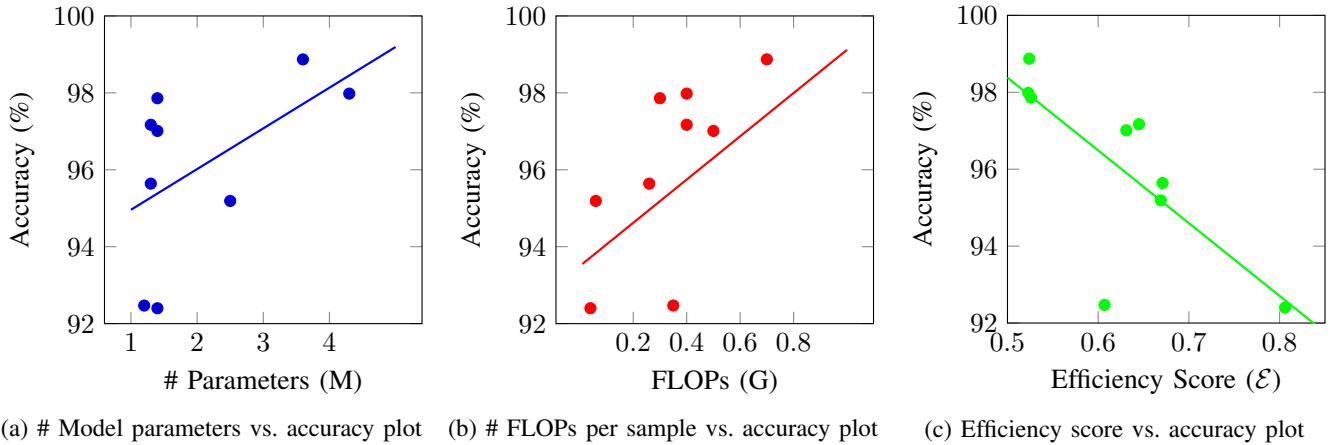
\begin{figure*}[t]
\caption{Accuracy comparison of LiMs across (a) parameters, (b) FLOPs per sample, and (c) efficiency score. Linear trend lines are shown with the corresponding $R^2$ values: (a) 0.27, (b) 0.33, and (c) 0.55.}
\resizebox{\textwidth}{!}{
\centering

\begin{minipage}{0.31\textwidth}
\centering
\begin{tikzpicture} 
\begin{axis}[
    xlabel={\# Parameters (M)},
    ylabel={Accuracy (\%)},
    ymin=92, ymax=100,
    xtick={1, 2, 3, 4},
    grid=none,
    width=\linewidth,
    height=5.5cm
]
\addplot+[only marks, mark=*, color=blue] coordinates {
    (1.4, 92.40) (1.3, 95.64) (2.5, 95.19) (1.3, 97.17)
    (1.4, 97.01) (1.2, 92.47) (1.4, 97.86) (3.6, 98.87) (4.3, 97.98)
};
\addplot [domain=1:5, color=blue, thick] {1.058 * x + 93.902};
\end{axis}
\end{tikzpicture}
\subcaption{\# Model parameters vs. accuracy plot}\label{fig:efficiency-trends-params}
\end{minipage}
\hfill

\begin{minipage}{0.31\textwidth}
\centering
\begin{tikzpicture}
\begin{axis}[
    xlabel={FLOPs (G)},
    ylabel={Accuracy (\%)},
    ymin=92, ymax=100,
    xtick={0.2, 0.4, 0.6, 0.8},
    grid=none,
    width=\linewidth,
    height=5.5cm
]
\addplot+[only marks, mark=*, color=red, mark options={fill=red}] coordinates {
    (0.04, 92.40) (0.26, 95.64) (0.06, 95.19) (0.40, 97.17)
    (0.50, 97.01) (0.35, 92.47) (0.30, 97.86) (0.70, 98.87) (0.40, 97.98)
};
\addplot [domain=0.01:1, color=red, thick] {5.624 * x + 93.495};
\end{axis}
\end{tikzpicture}
\subcaption{\# FLOPs per sample vs. accuracy plot}\label{fig:efficiency-trends-flops}
\end{minipage}
\hfill

\begin{minipage}{0.31\textwidth}
\centering
\begin{tikzpicture}
\begin{axis}[
    xlabel={Efficiency Score ($\mathcal{E}$)},
    ylabel={Accuracy (\%)},
    ymin=92, ymax=100,
    xmin=0.5, xmax=0.85,
    grid=none,
    width=\textwidth,
    height=5.5cm
]
\addplot+[only marks, mark=*, color=green, mark options={fill=green}] coordinates {
    (0.806, 92.40) (0.671, 95.64) (0.669, 95.19) (0.645, 97.17)
    (0.631, 97.01) (0.607, 92.47) (0.526, 97.86) (0.524, 98.87) (0.523, 97.98)
};
\addplot [domain=0.5:0.85, color=green, thick] {-18.92 * x + 107.84};
\end{axis}
\end{tikzpicture}
\subcaption{Efficiency score vs. accuracy plot}\label{fig:efficiency-trends-es}
\end{minipage}
}
\label{fig:efficiency-trends}
\end{figure*}

\subsection{Efficiency vs. Accuracy Tradeoff}
To better understand the tradeoff between efficiency and accuracy, we plot clean spatial testing accuracy against each LiM's parameter count, FLOPs per sample, and efficiency score as seen in Figure \ref{fig:efficiency-trends}. Clean spatial testing results are used because they provide a wider range of accuracy values compared to spatial testing which are all approximately 99\%. We omit MnasNet and the baseline methods from this analysis as Section \ref{subsec:results-clean} shows that MnasNet is unsuitable for AIGI detection and Section \ref{subsec:results-eff} highlights Lađević et al.'s excessive FLOPs per sample and SpottingDiffusion's disproportionate parameter count.

As shown in Figure \ref{fig:efficiency-trends-params}, parameter count and detection accuracy are weakly related ($R^2=0.27$), showing that heavily parametrized models are not universally effective. Instead, architectural design and features---such as attention mechanisms in LiViTs---provide more meaningful insight in model performance.

Figure~\ref{fig:efficiency-trends-flops} reveals a slightly stronger relationship between FLOPs and detection accuracy ($R^2=0.33$), where models with higher computational budgets tend to perform better. However, this trend is still modest and exceptions are evident with several models achieving high accuracy with relatively low FLOPs. This highlights the importance of how computation is structured rather than how much is used.

Finally, Figure~\ref{fig:efficiency-trends-es} shows a notably stronger correlation between our proposed efficiency score ($\mathcal{E}$) and detection accuracy ($R^2=0.55$). This relationship is negative however, indicating that the inverse of our efficiency score, would better capture the trade-off between model complexity and performance. It is also important to note that our $\mathcal{E}$-configuration (\(\lambda_1 =0.5\), \(\lambda_2 =0.25\), and \(\lambda_3 =0.25\)) represents one approach to model selection. Varying $\mathcal{E}$'s weighting coefficients can significantly influence selection outcomes and should be re-evaluated and validated on varied problem constraints

\section{Conclusion}\label{sec:conclustion}
In this paper, we present the first comprehensive evaluation of lightweight machine learning models (LiMs) for Artificial Intelligence-Generated Image (AIGI) detection. Through extensive benchmarking, we demonstrate that LiMs can achieve high detection accuracy (exceeding 99\%) while maintaining extremely low computational overheads (fewer than 10 million parameters and under 1 GFLOPs per sample).Our results reinforce prior work and further confirm that the frequency domain of AIGI images contains detectable spectral artifacts that can enhance classification performance. While larger models often perform better, our results indicate that architectural design plays a more critical role in determining success. Additionally, we find that LiMs exhibit strong resilience to basic adversarial perturbations in spatial and fusion-based settings, highlighting their potential as robust, deployable solutions.

Looking ahead, there are several potential research directions. Firstly, investigating more lightweight architectures with across diverse datasets would help assess the generalizability of LiMs in AIGI detection. Secondly, expanding the scope of detection tasks to include generator attribution and cross-generator generalization would also provide alternate problem domains for LiMs. Thirdly, we encourage the exploration of alternative model selection strategies, the use of knowledge distillation, and neural architecture search (NAS) to provide different perspectives on how architectural compactness impacts detection capacity. Finally, as discussed in Section \ref{subsec:results-attack}, further exploration of adversarial attacks and corresponding defenses in lightweight detection settings is warranted.



\begin{thebibliography}{10}
\providecommand{\url}[1]{#1}
\csname url@samestyle\endcsname
\providecommand{\newblock}{\relax}
\providecommand{\bibinfo}[2]{#2}
\providecommand{\BIBentrySTDinterwordspacing}{\spaceskip=0pt\relax}
\providecommand{\BIBentryALTinterwordstretchfactor}{4}
\providecommand{\BIBentryALTinterwordspacing}{\spaceskip=\fontdimen2\font plus
\BIBentryALTinterwordstretchfactor\fontdimen3\font minus \fontdimen4\font\relax}
\providecommand{\BIBforeignlanguage}[2]{{%
\expandafter\ifx\csname l@#1\endcsname\relax
\typeout{** WARNING: IEEEtran.bst: No hyphenation pattern has been}%
\typeout{** loaded for the language `#1'. Using the pattern for}%
\typeout{** the default language instead.}%
\else
\language=\csname l@#1\endcsname
\fi
#2}}
\providecommand{\BIBdecl}{\relax}
\BIBdecl

\bibitem{aigcconcern-lu2023}
Z.~Lu, D.~Huang, L.~Bai, J.~Qu, C.~Wu, X.~Liu, and W.~Ouyang, ``Seeing is not always believing: Benchmarking human and model perception of ai-generated images,'' pp. 25\,435--25\,447, 2023.

\bibitem{aigcconcern-peng2024}
Q.~Peng, Y.~Lu, Y.~Peng, S.~Qian, X.~Liu, and C.~Shen, ``Crafting synthetic realities: Examining visual realism and misinformation potential of photorealistic ai-generated images,'' pp. 1--12, 2025.

\bibitem{aigcconcern-misinforeview2024}
R.~DiResta and J.~A. Goldstein, ``How spammers and scammers leverage ai-generated images on facebook for audience growth,'' 2024.

\bibitem{smstats-tsvetkova2023}
R.~Tsvetkova, ``99 amazing social media statistics and facts,'' https://www.brandwatch.com/blog/amazing-social-media-statistics-and-facts/, 2023.

\bibitem{smstats-qut2023}
P.~Dootson, ``3.2 billion images and 720,000 hours of video are shared online daily. can you sort real from fake?'' https://www.qut.edu.au/insights/business/3.2-billion-images-and-720000-hours-of-video-are-shared-online-daily.-can-you-sort-real-from-fake, 2021, {Queensland University of Technology}.

\bibitem{smstats-broz2024photo}
M.~Broz, ``Photo statistics: How many photos are taken every day?'' https://photutorial.com/photos-statistics/, 2025.

\bibitem{smstats-clegg2024}
N.~Clegg, ``Labeling ai-generated images on facebook, instagram and threads,'' https://about.fb.com/news/2024/02/labeling-ai-generated-images-on-facebook-instagram-and-threads/, 2024.

\bibitem{smstats-tiktok2023}
{TikTok Newsroom}, ``New labels for disclosing ai-generated content,'' https://newsroom.tiktok.com/en-us/new-labels-for-disclosing-ai-generated-content, 2023.

\bibitem{smstats-xauthenticity2025}
{X Help Center}, ``Authenticity,'' https://help.x.com/en/rules-and-policies/authenticity, 2025.

\bibitem{related-guo2025}
M.~Guo, Y.~Hu, Z.~Jiang, Z.~Li, A.~Sadovnik, A.~Daw, and N.~Gong, ``Ai-generated image detection: Passive or watermark?'' 2024.

\bibitem{related-durall2020}
R.~Durall, M.~Keuper, and J.~Keuper, ``Watch your up-convolution: Cnn based generative deep neural networks are failing to reproduce spectral distributions,'' in \emph{Proceedings of the IEEE/CVF conference on computer vision and pattern recognition}, 2020, pp. 7890--7899.

\bibitem{related-tomen2021}
N.~Tomen and J.~C. van Gemert, ``Spectral leakage and rethinking the kernel size in cnns,'' in \emph{Proceedings of the IEEE/CVF International Conference on Computer Vision}, 2021, pp. 5138--5147.

\bibitem{related-he2021}
Y.~He, N.~Yu, M.~Keuper, and M.~Fritz, ``Beyond the spectrum: Detecting deepfakes via re-synthesis,'' 2021.

\bibitem{related-jeong2022}
Y.~Jeong, D.~Kim, S.~Min, S.~Joe, Y.~Gwon, and J.~Choi, ``Bihpf: Bilateral high-pass filters for robust deepfake detection,'' in \emph{Proceedings of the IEEE/CVF Winter Conference on Applications of Computer Vision}, 2022, pp. 48--57.

\bibitem{related-deng2024}
J.~Deng, C.~Lin, Z.~Zhao, S.~Liu, Q.~Wang, and C.~Shen, ``A survey of defenses against ai-generated visual media: Detection, disruption, and authentication,'' 2024.

\bibitem{related-Kassis2024}
A.~Kassis and U.~Hengartner, ``Unmarker: A universal attack on defensive watermarking,'' \emph{arXiv preprint arXiv:2405.08363}, 2024.

\bibitem{related_Zhu2018}
J.~Zhu, R.~Kaplan, J.~Johnson, and L.~Fei-Fei, ``Hidden: Hiding data with deep networks,'' in \emph{Proceedings of the European conference on computer vision (ECCV)}, 2018, pp. 657--672.

\bibitem{related-fernandez2023}
P.~Fernandez, G.~Couairon, H.~J{\'e}gou, M.~Douze, and T.~Furon, ``The stable signature: Rooting watermarks in latent diffusion models,'' in \emph{Proceedings of the IEEE/CVF International Conference on Computer Vision}, 2023, pp. 22\,466--22\,477.

\bibitem{related-kim2024}
C.~Kim, K.~Min, M.~Patel, S.~Cheng, and Y.~Yang, ``Wouaf: Weight modulation for user attribution and fingerprinting in text-to-image diffusion models,'' in \emph{Proceedings of the IEEE/CVF Conference on Computer Vision and Pattern Recognition}, 2024, pp. 8974--8983.

\bibitem{related-yang2024}
Z.~Yang, K.~Zeng, K.~Chen, H.~Fang, W.~Zhang, and N.~Yu, ``Gaussian shading: Provable performance-lossless image watermarking for diffusion models,'' in \emph{Proceedings of the IEEE/CVF Conference on Computer Vision and Pattern Recognition}, 2024, pp. 12\,162--12\,171.

\bibitem{related-wen2023}
Y.~Wen, J.~Kirchenbauer, J.~Geiping, and T.~Goldstein, ``Tree-rings watermarks: Invisible fingerprints for diffusion images,'' vol.~36, 2023, pp. 58\,047--58\,063.

\bibitem{related-wang2020}
R.~Wang, F.~Juefei-Xu, Q.~Guo, Y.~Huang, X.~Xie, L.~Ma, Y.~Liu, and L.~Wang, ``Deeptag: Robust image tagging for deepfake provenance. arxiv preprint arxiv,'' 2020.

\bibitem{related-tancik2020}
M.~Tancik, B.~Mildenhall, and R.~Ng, ``Stegastamp: Invisible hyperlinks in physical photographs,'' in \emph{Proceedings of the IEEE/CVF conference on computer vision and pattern recognition}, 2020, pp. 2117--2126.

\bibitem{related-Rossler2019}
A.~Rossler, D.~Cozzolino, L.~Verdoliva, C.~Riess, J.~Thies, and M.~Nie{\ss}ner, ``Faceforensics++: Learning to detect manipulated facial images,'' in \emph{Proceedings of the IEEE/CVF international conference on computer vision}, 2019, pp. 1--11.

\bibitem{related-wang2020-CNNSpot}
S.-Y. Wang, O.~Wang, R.~Zhang, A.~Owens, and A.~A. Efros, ``Cnn-generated images are surprisingly easy to spot... for now,'' in \emph{Proceedings of the IEEE/CVF conference on computer vision and pattern recognition}, 2020, pp. 8695--8704.

\bibitem{related-tan2023}
C.~Tan, Y.~Zhao, S.~Wei, G.~Gu, and Y.~Wei, ``Learning on gradients: Generalized artifacts representation for gan-generated images detection,'' in \emph{Proceedings of the IEEE/CVF Conference on Computer Vision and Pattern Recognition}, 2023, pp. 12\,105--12\,114.

\bibitem{related-ojha2023}
U.~Ojha, Y.~Li, and Y.~J. Lee, ``Towards universal fake image detectors that generalize across generative models,'' in \emph{Proceedings of the IEEE/CVF Conference on Computer Vision and Pattern Recognition}, 2023, pp. 24\,480--24\,489.

\bibitem{related-wang2023}
Z.~Wang, J.~Bao, W.~Zhou, W.~Wang, H.~Hu, H.~Chen, and H.~Li, ``Dire for diffusion-generated image detection,'' in \emph{Proceedings of the IEEE/CVF International Conference on Computer Vision}, 2023, pp. 22\,445--22\,455.

\bibitem{related-zhong2024}
N.~Zhong, Y.~Xu, S.~Li, Z.~Qian, and X.~Zhang, ``Patchcraft: Exploring texture patch for efficient ai-generated image detection,'' \emph{arXiv preprint arXiv:2311.12397}, 2023.

\bibitem{related-frank2020}
J.~Frank, T.~Eisenhofer, L.~Sch{\"o}nherr, A.~Fischer, D.~Kolossa, and T.~Holz, ``Leveraging frequency analysis for deep fake image recognition,'' in \emph{International conference on machine learning}.\hskip 1em plus 0.5em minus 0.4em\relax PMLR, 2020, pp. 3247--3258.

\bibitem{related-yu2019}
N.~Yu, L.~S. Davis, and M.~Fritz, ``Attributing fake images to gans: Learning and analyzing gan fingerprints,'' in \emph{Proceedings of the IEEE/CVF international conference on computer vision}, 2019, pp. 7556--7566.

\bibitem{related-yan2025}
S.~Yan, O.~Li, J.~Cai, Y.~Hao, X.~Jiang, Y.~Hu, and W.~Xie, ``A sanity check for ai-generated image detection,'' \emph{arXiv preprint arXiv:2406.19435}, 2024.

\bibitem{related-sha2023}
Z.~Sha, Z.~Li, N.~Yu, and Y.~Zhang, ``De-fake: Detection and attribution of fake images generated by text-to-image generation models,'' in \emph{Proceedings of the 2023 ACM SIGSAC conference on computer and communications security}, 2023, pp. 3418--3432.

\bibitem{related-radford2021}
A.~Radford, J.~W. Kim, C.~Hallacy, A.~Ramesh, G.~Goh, S.~Agarwal, G.~Sastry, A.~Askell, P.~Mishkin, J.~Clark \emph{et~al.}, ``Learning transferable visual models from natural language supervision,'' in \emph{International conference on machine learning}.\hskip 1em plus 0.5em minus 0.4em\relax PmLR, 2021, pp. 8748--8763.

\bibitem{related-Szegedy2016}
C.~Szegedy, V.~Vanhoucke, S.~Ioffe, J.~Shlens, and Z.~Wojna, ``Rethinking the inception architecture for computer vision,'' in \emph{Proceedings of the IEEE conference on computer vision and pattern recognition}, 2016, pp. 2818--2826.

\bibitem{related-dosovitskiy2020}
A.~Dosovitskiy, L.~Beyer, A.~Kolesnikov, D.~Weissenborn, X.~Zhai, T.~Unterthiner, M.~Dehghani, M.~Minderer, G.~Heigold, S.~Gelly \emph{et~al.}, ``An image is worth 16x16 words: Transformers for image recognition at scale,'' \emph{arXiv preprint arXiv:2010.11929}, 2020.

\bibitem{related-wang2021-fakespotter}
R.~Wang, F.~Juefei-Xu, L.~Ma, X.~Xie, Y.~Huang, J.~Wang, and Y.~Liu, ``Fakespotter: A simple yet robust baseline for spotting ai-synthesized fake faces,'' 2019.

\bibitem{related-VGGFace}
R.~C. Malli, ``keras-vggface: Vggface implementation with keras framework,'' https://github.com/rcmalli/keras-vggface, 2017.

\bibitem{related-ladevic2024}
A.~L. La{\dj}evi{\'c}, T.~Kramberger, R.~Kramberger, and D.~Vlahek, ``Detection of ai-generated synthetic images with a lightweight cnn,'' \emph{AI}, vol.~5, no.~3, p. 1575, 2024.

\bibitem{related-mulki2024}
M.~Mulki and S.~Mulki, ``Spottingdiffusion:re using transfer learning to detect latent diffusion model-synthesized images,'' \emph{Journal of Emerging Investigators}, 01 2024.

\bibitem{related-rombach2022}
R.~Rombach, A.~Blattmann, D.~Lorenz, P.~Esser, and B.~Ommer, ``High-resolution image synthesis with latent diffusion models,'' in \emph{Proceedings of the IEEE/CVF conference on computer vision and pattern recognition}, 2022, pp. 10\,684--10\,695.

\bibitem{related-midjourney}
{MidJourney, Inc.}, ``Midjourney,'' https://www.midjourney.com/, 2022.

\bibitem{related-ramesh2021}
A.~Ramesh, M.~Pavlov, G.~Goh, S.~Gray, C.~Voss, A.~Radford, M.~Chen, and I.~Sutskever, ``Zero-shot text-to-image generation,'' in \emph{International conference on machine learning}.\hskip 1em plus 0.5em minus 0.4em\relax Pmlr, 2021, pp. 8821--8831.

\bibitem{related-zhu2023}
M.~Zhu, H.~Chen, Q.~Yan, X.~Huang, G.~Lin, W.~Li, Z.~Tu, H.~Hu, J.~Hu, and Y.~Wang, ``Genimage: A million-scale benchmark for detecting ai-generated image,'' \emph{Advances in Neural Information Processing Systems}, vol.~36, pp. 77\,771--77\,782, 2023.

\bibitem{related-dhariwal2021}
P.~Dhariwal and A.~Nichol, ``Diffusion models beat gans on image synthesis,'' \emph{Advances in neural information processing systems}, vol.~34, pp. 8780--8794, 2021.

\bibitem{related-nichol2021}
A.~Nichol, P.~Dhariwal, A.~Ramesh, P.~Shyam, P.~Mishkin, B.~McGrew, I.~Sutskever, and M.~Chen, ``Glide: Towards photorealistic image generation and editing with text-guided diffusion models,'' \emph{arXiv preprint arXiv:2112.10741}, 2021.

\bibitem{related-wukong2022}
{MindSpore ModelZoo}, ``Wukong,'' https://xihe.mindspore.cn/modelzoo/wukong, 2022.

\bibitem{related-Egiazarian2024}
V.~Egiazarian, D.~Kuznedelev, A.~Voronov, R.~Svirschevski, M.~Goin, D.~Pavlov, D.~Alistarh, and D.~Baranchuk, ``Accurate compression of text-to-image diffu-sion models via vector quantization,'' \emph{arXiv preprint arXiv:2409.00492}, 2024.

\bibitem{related-brock2018}
A.~Brock, J.~Donahue, and K.~Simonyan, ``Large scale gan training for high fidelity natural image synthesis,'' \emph{arXiv preprint arXiv:1809.11096}, 2018.

\bibitem{related-pytorchzoo}
``Models and pre-trained weights,'' https://pytorch.org/vision/stable/models.html.

\bibitem{related-timm}
R.~Wightman, ``Pytorch image models,'' https://github.com/rwightman/pytorch-image-models, 2019.

\bibitem{related-Krizhevsky2012}
A.~Krizhevsky, I.~Sutskever, and G.~E. Hinton, ``Imagenet classification with deep convolutional neural networks,'' vol.~60, no.~6.\hskip 1em plus 0.5em minus 0.4em\relax AcM New York, NY, USA, 2017, pp. 84--90.

\bibitem{related-huang2017}
G.~Huang, Z.~Liu, L.~Van Der~Maaten, and K.~Q. Weinberger, ``Densely connected convolutional networks,'' in \emph{Proceedings of the IEEE conference on computer vision and pattern recognition}, 2017, pp. 4700--4708.

\bibitem{related-tan2021}
M.~Tan and Q.~Le, ``Efficientnetv2: Smaller models and faster training,'' in \emph{International conference on machine learning}.\hskip 1em plus 0.5em minus 0.4em\relax PMLR, 2021, pp. 10\,096--10\,106.

\bibitem{related-szegedy2015}
C.~Szegedy, W.~Liu, Y.~Jia, P.~Sermanet, S.~Reed, D.~Anguelov, D.~Erhan, V.~Vanhoucke, and A.~Rabinovich, ``Going deeper with convolutions,'' in \emph{Proceedings of the IEEE conference on computer vision and pattern recognition}, 2015, pp. 1--9.

\bibitem{related-tan2019}
M.~Tan, B.~Chen, R.~Pang, V.~Vasudevan, M.~Sandler, A.~Howard, and Q.~V. Le, ``Mnasnet: Platform-aware neural architecture search for mobile,'' in \emph{Proceedings of the IEEE/CVF conference on computer vision and pattern recognition}, 2019, pp. 2820--2828.

\bibitem{related-sandler2018}
M.~Sandler, A.~Howard, M.~Zhu, A.~Zhmoginov, and L.-C. Chen, ``Mobilenetv2: Inverted residuals and linear bottlenecks,'' in \emph{Proceedings of the IEEE conference on computer vision and pattern recognition}, 2018, pp. 4510--4520.

\bibitem{related-howard2019}
A.~Howard, M.~Sandler, G.~Chu, L.-C. Chen, B.~Chen, M.~Tan, W.~Wang, Y.~Zhu, R.~Pang, V.~Vasudevan \emph{et~al.}, ``Searching for mobilenetv3,'' in \emph{Proceedings of the IEEE/CVF international conference on computer vision}, 2019, pp. 1314--1324.

\bibitem{related-xu2022}
J.~Xu, Y.~Pan, X.~Pan, S.~Hoi, Z.~Yi, and Z.~Xu, ``Regnet: self-regulated network for image classification,'' \emph{IEEE Transactions on Neural Networks and Learning Systems}, vol.~34, no.~11, pp. 9562--9567, 2022.

\bibitem{related-zhang2018}
X.~Zhang, X.~Zhou, M.~Lin, and J.~Sun, ``Shufflenet: An extremely efficient convolutional neural network for mobile devices,'' in \emph{Proceedings of the IEEE conference on computer vision and pattern recognition}, 2018, pp. 6848--6856.

\bibitem{related-iandola2016}
F.~N. Iandola, S.~Han, M.~W. Moskewicz, K.~Ashraf, W.~J. Dally, and K.~Keutzer, ``Squeezenet: Alexnet-level accuracy with 50x fewer parameters and< 0.5 mb model size,'' \emph{arXiv preprint arXiv:1602.07360}, 2016.

\bibitem{related-xu2021}
W.~Xu, Y.~Xu, T.~Chang, and Z.~Tu, ``Co-scale conv-attentional image transformers,'' in \emph{Proceedings of the IEEE/CVF international conference on computer vision}, 2021, pp. 9981--9990.

\bibitem{related-maaz2022}
M.~Maaz, A.~Shaker, H.~Cholakkal, S.~Khan, S.~W. Zamir, R.~M. Anwer, and F.~Shahbaz~Khan, ``Edgenext: efficiently amalgamated cnn-transformer architecture for mobile vision applications,'' in \emph{European conference on computer vision}.\hskip 1em plus 0.5em minus 0.4em\relax Springer, 2022, pp. 3--20.

\bibitem{related-li2022}
Y.~Li, G.~Yuan, Y.~Wen, J.~Hu, G.~Evangelidis, S.~Tulyakov, Y.~Wang, and J.~Ren, ``Efficientformer: Vision transformers at mobilenet speed,'' \emph{Advances in Neural Information Processing Systems}, vol.~35, pp. 12\,934--12\,949, 2022.

\bibitem{related-vasu2023}
P.~K.~A. Vasu, J.~Gabriel, J.~Zhu, O.~Tuzel, and A.~Ranjan, ``Fastvit: A fast hybrid vision transformer using structural reparameterization,'' in \emph{Proceedings of the IEEE/CVF international conference on computer vision}, 2023, pp. 5785--5795.

\bibitem{related-graham2021}
B.~Graham, A.~El-Nouby, H.~Touvron, P.~Stock, A.~Joulin, H.~J{\'e}gou, and M.~Douze, ``Levit: a vision transformer in convnet's clothing for faster inference,'' in \emph{Proceedings of the IEEE/CVF international conference on computer vision}, 2021, pp. 12\,259--12\,269.

\bibitem{related-mehta2021}
S.~Mehta and M.~Rastegari, ``Mobilevit: light-weight, general-purpose, and mobile-friendly vision transformer,'' \emph{arXiv preprint arXiv:2110.02178}, 2021.

\bibitem{related-mehta2022}
------, ``Separable self-attention for mobile vision transformers,'' \emph{arXiv preprint arXiv:2206.02680}, 2022.

\end{thebibliography}
\end{document}